\documentclass[10pt,twocolumn,letterpaper]{article}

\usepackage{iccv}              % To produce the CAMERA-READY version
\definecolor{iccvblue}{rgb}{0.21,0.49,0.74}
\usepackage[pagebackref,breaklinks,colorlinks,allcolors=iccvblue]{hyperref}
\usepackage{bbding}
\usepackage{tcolorbox}
\usepackage{tabularx}
\usepackage{multirow}
\usepackage{graphicx}

\def\paperID{3349} % *** Enter the Paper ID here
\def\confName{ICCV}
\def\confYear{2025}

\title{IGD: Instructional Graphic Design with Multimodal Layer Generation}

\author{Yadong Qu$^1$\thanks{Work done during the internship at YuanShi Technology.}, Shancheng Fang$^2$\thanks{Corresponding author.}, Yuxin Wang$^2$,\\
Xiaorui Wang$^2$, Zhineng Chen$^3$, Hongtao Xie$^1$, Yongdong Zhang$^1$\\
$^1$University of Science and Technology of China $^2$YuanShi Technology \\
$^3$Institute of Trustworthy Embodied AI, Fudan University \\
{\tt\small qqqyd@mail.ustc.edu.cn, \{fangsc, wangyx58, htxie, zhyd73\}@ustc.edu.cn, harrywxr@outlook.com}
}

\begin{document}
\maketitle
\begin{abstract}
Graphic design visually conveys information and data by creating and combining text, images and graphics. Two-stage methods that rely primarily on layout generation lack creativity and intelligence, making graphic design still labor-intensive. Existing diffusion-based methods generate non-editable graphic design files at image level with poor legibility in visual text rendering, which prevents them from achieving satisfactory and practical automated graphic design. In this paper, we propose Instructional Graphic Designer (IGD) to swiftly generate multimodal layers with editable flexibility with only natural language instructions. IGD adopts a new paradigm that leverages parametric rendering and image asset generation. First, we develop a design platform and establish a standardized format for multi-scenario design files, thus laying the foundation for scaling up data. Second, IGD utilizes the multimodal understanding and reasoning capabilities of MLLM to accomplish attribute prediction, sequencing and layout of layers. It also employs a diffusion model to generate image content for assets. By enabling end-to-end training, IGD architecturally supports scalability and extensibility in complex graphic design tasks. The superior experimental results demonstrate that IGD offers a new solution for graphic design.

\end{abstract}    
\section{Introduction}
\label{sec:intro}

\begin{figure}[htbp]
    \centering
    \includegraphics[width=\linewidth]{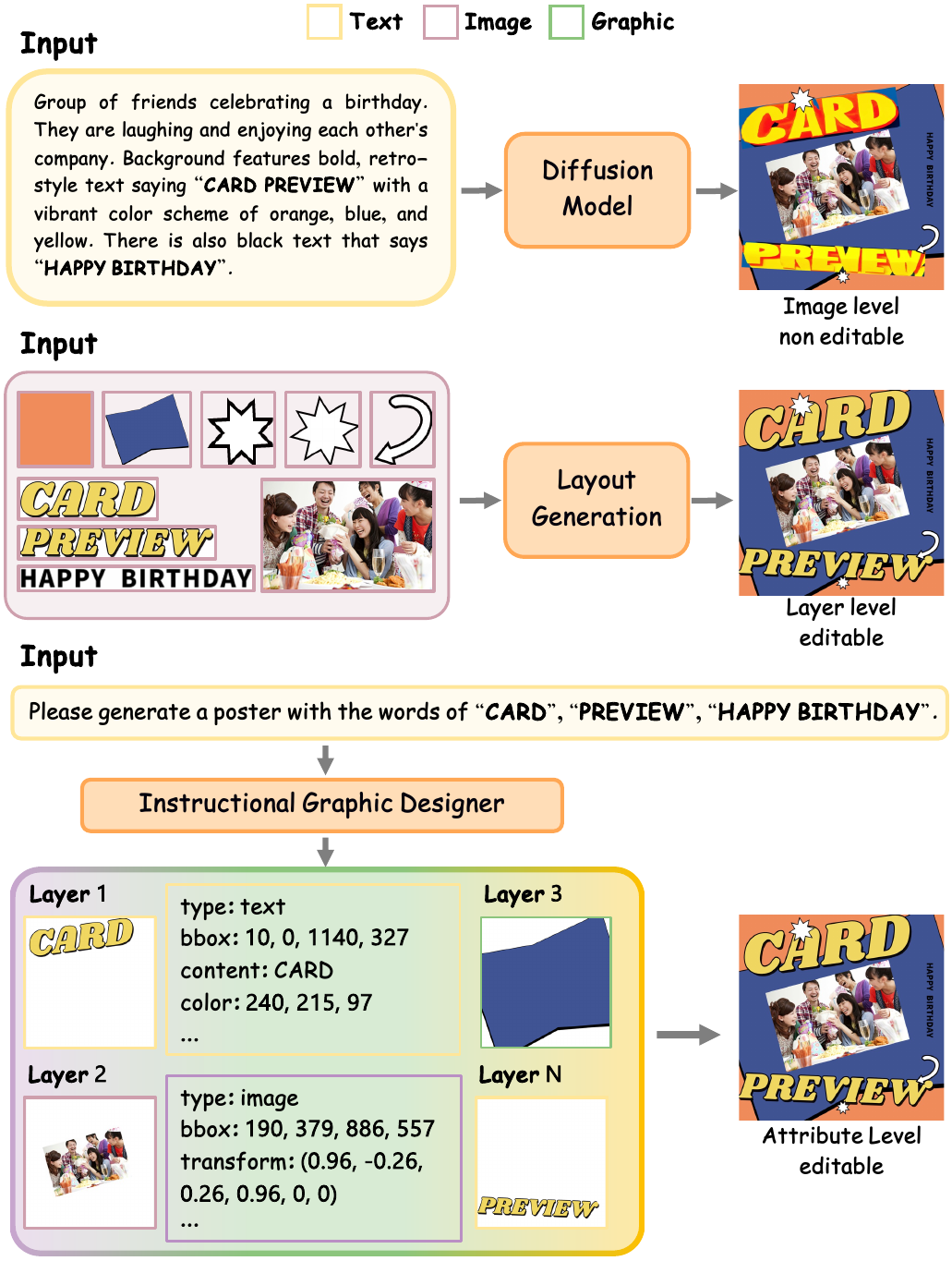}
    \caption{Method comparison. Diffusion models struggle with legible text and lack editing flexibility. Layout generation requires pre-designed elements, limiting creativity. IGD generates layered content and layouts solely from instructions.}
    \label{vis:fig1}
\end{figure}

With the rapid advancements of deep learning technology and computational power, scaling laws have driven revolutionary success in recent years. In natural language processing tasks, large language models (LLMs)~\cite{gpt4, llama, vicuna, qwen} have demonstrated remarkable capabilities, while diffusion models~\cite{ddpm, stablediffusion, imagen, dalle3, anytext} have also achieved impressive results in text-to-image generation, making previously challenging tasks achievable. This success owes much to scaling laws. Graphic design, as a fundamental task in conveying visual and textual information, traditionally requires substantial expertise and human effort to create aesthetically pleasing works. However, there is no sufficiently intelligent and effective methods in this multimodal scenario to create layer editable graphic design files based on instructions.

Previous methods attempted to achieve it in a straightforward way using two separated stages. In the first stage, all required text, image and graphic assets are obtained by any means, such as human selection, retrieval. Subsequently, layout generation methods~\cite{cglgan, layoutgpt, posterllava, graphist} are used to predict the position and size of all assets in image form, creating an aesthetically pleasing visual layout. However, they are subject to some limitations due to the requirement of predefined elements. On the one hand, users are responsible for collecting or designing all elements. The lack of model creativity demands a considerable amount of expertise and manual effort, detracting from the automation. On the other hand, predetermined design elements can constrain the typographic layout potential of the design. Discordant visual elements provided by inexperienced users can make the overall visual presentation lack aesthetic appeal.

Another feasible approach is to use text-to-image diffusion models~\cite{textdiffuser, glyphcontrol, glyphbyt5, anytext}, which can follow instructions and complete the task in a single stage. However, they still face substantial challenges: 1) Poor text rendering: The generated text often exhibits errors, omissions, or extraneous characters. This issue is particularly pronounced with small or dense text, where readability is severely compromised. 2) Limited flexibility: graphic design files are generated as flat images, making them non-editable and restricting users from customizing or refining the elements. 3) High threshold for use: Diffusion models demand detailed prompts for satisfactory results and different models respond variably to prompts, increasing complexity for users, especially novices, thus limiting their practical applicability. These challenges highlight the need for more user-friendly and adaptable solutions.

To improve the legibility of visual text while making the model intelligent enough to follow instructions and generate interactively editable graphic design files, we propose \textbf{I}nstructional \textbf{G}raphic \textbf{D}esigner. IGD is expected to acquire knowledge from massive data and generate multimodal layers in a single stage, including text, image and graphic layers. On the data aspect, we develop a design platform based on Penpot\footnote{\url{https://github.com/penpot/penpot}} to achieve rendering and interactive editing of multimodal layers. Specifically, a standardized format is established that allows graphic design files from various sources to be converted into a uniform format, thus enabling IGD to accomplish multi-scenario graphic design. On the model aspect, unlike prior approaches that primarily rely on diffusion models for graphic design, we adopt a parameter-prediction and rendering strategy. Specifically, we construct a multimodal comprehension and creation model by integrating the multimodal large language model (MLLM) with diffusion model to adapt to multimodal layers. The MLLM is adept at incorporating diverse data modalities and can effectively handle attribute prediction, sequencing, and arrange of layers with its extensive prior knowledge, multimodal understanding, and logical reasoning capbilities. Meanwhile, the diffusion model creatively generates image content, completing the visual elements of the design. The two work in synergy, optimized through end-to-end training, ensuring the generated image aligns harmoniously with other layers. Both the MLLM and diffusion models are highly scalable, and with our standardized data processing, the entire pipeline is optimized for scalability.

Compared to layout generation methods, IGD enables instruction-guided graphic design generation with creativity. By accomplishing both asset creation and layout arrangement within a single model, it not only significantly reduces workload but also enhances harmony between multimodal layers. Compared to diffusion-based methods, IGD avoids issues with unclear and illegible visual text and provides attribute-level editing flexibility, which greatly improves practicality and offers designers an inspiring tool to facilite their creative process. IGD is the first universal graphic design model to generate editable multimodal layers. Overall, our contributions are summarized as follows:
\begin{itemize}
    \item{Based on the Penpot platform, we propose the standardized format of multi-scenario graphic design files, thereby providing a solid foundation for data scaling and a multi-purpose universal graphic design method.}
    \item{The model adopts a composite structure of MLLM and diffusion model, supporting instruction based multimodal layer generation through data-driven end-to-end training.}
    \item{Qualitative and quantitative experimental results indicate that IGD can effectively generate multi-scenario graphic design files while maintaining the coherence and editability across all multimodal layers, offering a new and more versatile solution for graphic design.}
\end{itemize}

\section{Related Work}
\label{sec:related_work}

\subsection{Multimodal Large Language Model}
In recent years, reseachers have achieved remarkable process in large language models (LLMs)~\cite{gpt4, llama, vicuna, qwen}, pushing natural language processing performance to new heights across various tasks. By mapping the visual features extracted from the foundational vision model into the semantic space of LLMs, visual language models~\cite{gpt4v, flamingo, blip2, mplugowl, minigpt4, qwenvl} effectively integrate visual and language representations and achieve multimodal understanding. Furthermore, researchers have explored general MLLMs, which are designed to possess both multimodal understanding and creation capabilities. Some approaches~\cite{showo, seed, janus, emu3, chameleon} employ a VQ tokenizer~\cite{vqvae} to convert images into discrete tokens, integrating visual information by extending the vocabulary of LLMs. Other methods~\cite{dreamllm, seedx, transfusion, emu} operate on the continuous features of images, generating high-quality images by regressing image output features or applying end-to-end image-level supervision.

\subsection{Graphic Layout Generation}
Several methods have been explored for the layout generation task in two-stage graphic design. CGL-GAN~\cite{cglgan} and Autoposter~\cite{autoposter} use transformer models to generate layouts by understanding the semantic information of images and considering the spatial relationships between elements. Thanks to the logical reasoning capabilities of MLLMs, LayoutGPT~\cite{layoutgpt} guides the LLM to output layout information in the desired format through carefully designed structured prompts. PosterLLaVa~\cite{posterllava} employs MLLM to accomplish layout generation while meeting additional user requirements. Graphist~\cite{graphist}, based on traditional layout generation, introduces the hierarchical layout generation task, achieving reasonable layouts when the elements are input in an unordered manner. However, these methods lack the ability to create assets, requiring all multimodal layers to be pre-designed, which greatly increases labor costs and usage complexity. Additionally, pre-designed assets also contraint the potential visual performance of the design image.

\subsection{Visual Text Generation}
Diffusion models~\cite{ddpm, ddim, stablediffusion, controlnet} are a class of generative models that create data by iteratively refining noise through a denoising process. However, visual text generated by diffusion models often contains errors, omissions, or distorted characters. To address this issue, some methods~\cite{imagen, ediffi, liu2022character} find that using T5-based text encoders~\cite{t5} can effectively improve the quality of visual text rendering. On the other hand, ControlNet-based methods~\cite{controlnet, glyphdraw, glyphcontrol, glyphbyt5, anytext, textgen} leverage glyph images of the target text presented in a standard font as conditions, significantly enhancing the quality of visual text rendering. However, these methods still perform poorly in cases involving small or dense text, leading to generated text that is difficult for humans to read or unidentifiable by text recognition models~\cite{du2022@svtr, zheng2024cdistnet, visu, du2025instruction}. Moreover, they are not applicable to other scenarios such as slide generation. COLE~\cite{cole} generates background and foreground images separately using diffusion models, and then employs a large language model to predict text attributes. Although COLE can intelligently follow instructions, it imposes significant restrictions on the design flexibility, such as only including the two images located at the bottom layers and offering limited display effects. In contrast, our method represents a more general, flexible, and practical design pipeline that is closer to practical applications.

\begin{figure}[htbp]
    \centering
    \includegraphics[width=0.9\linewidth]{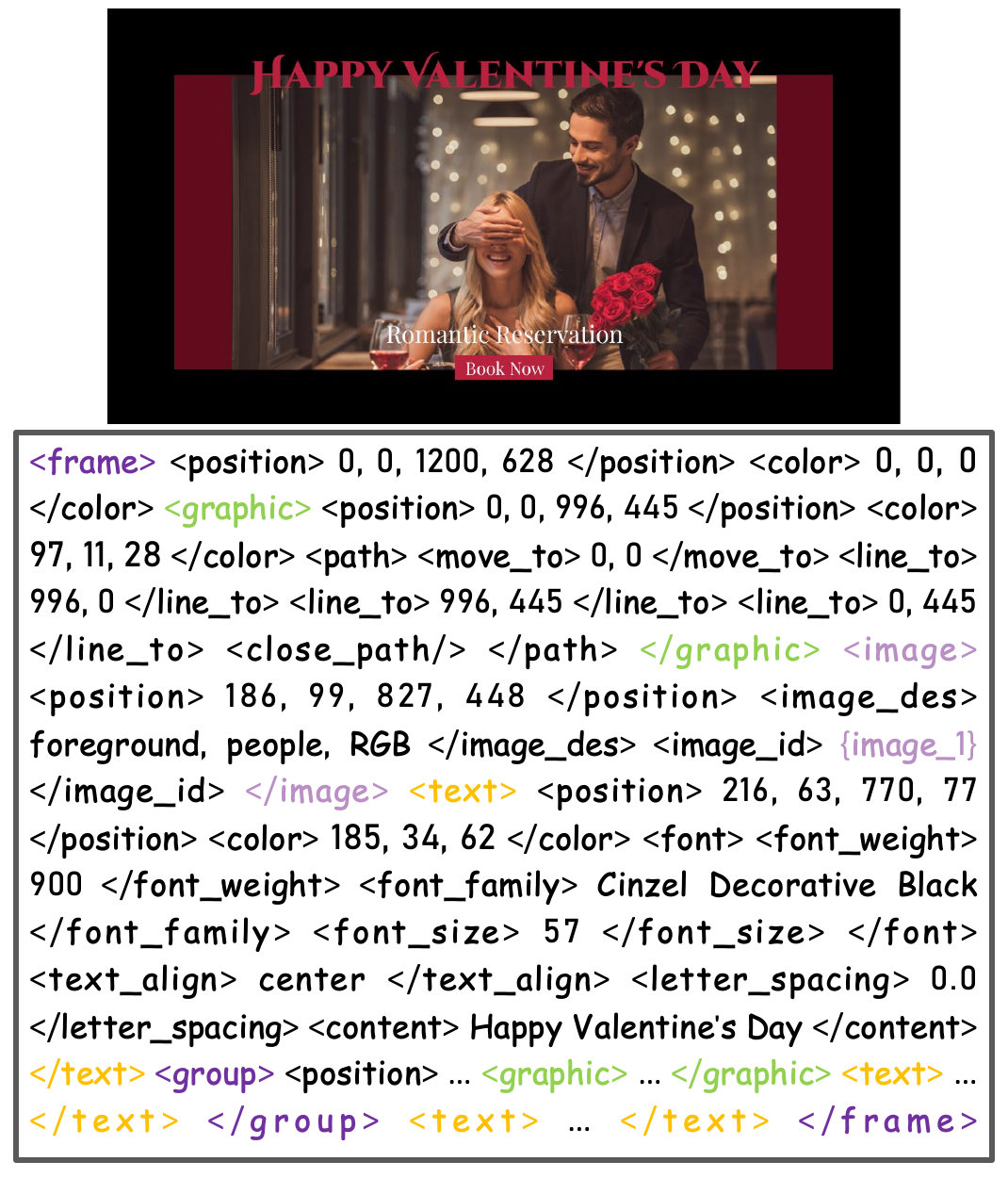}
    \caption{An example of representing a design file in the standardized format.}
    \label{vis:stand_format}
\end{figure}

\section{Methodology}
\subsection{Data Perspective}
\subsubsection{Standardized Format}
Diffusion methods rely on data composed of text-image pairs, while layout generation methods use data in which all layers are presented as images, containing only limited attribute information. The absence of key attributes makes it difficult to serve IGD. Graphic design files consist of a series of multimodal layers combined according to their relative positions and stacking order. Due to the interwoven nature of multimodal data and structural variations across different design file types, we adopt Penpot, the largest open-source design platform, for comprehensive data collection, storage and annotation. It plays a crucial role in real-time rendering, providing editing flexibility and enabling data scalability. As shown in \cref{vis:stand_format}, multimodal layers are categorized into frame, group, graphic, text and image. To unify these multimodal layers, we convert each layer type into a standardized sequence of special tokens and text (detailed attributes of each layer type are described in the supplementary materials). This LLM-friendly standardization enables a consistent representation of graphic design files across different scenarios, laying the foundation for data scalability and equipping the model with the ability to generate design files across diverse contexts.

\begin{figure*}[htbp]
    \centering
    \includegraphics[width=0.95\textwidth]{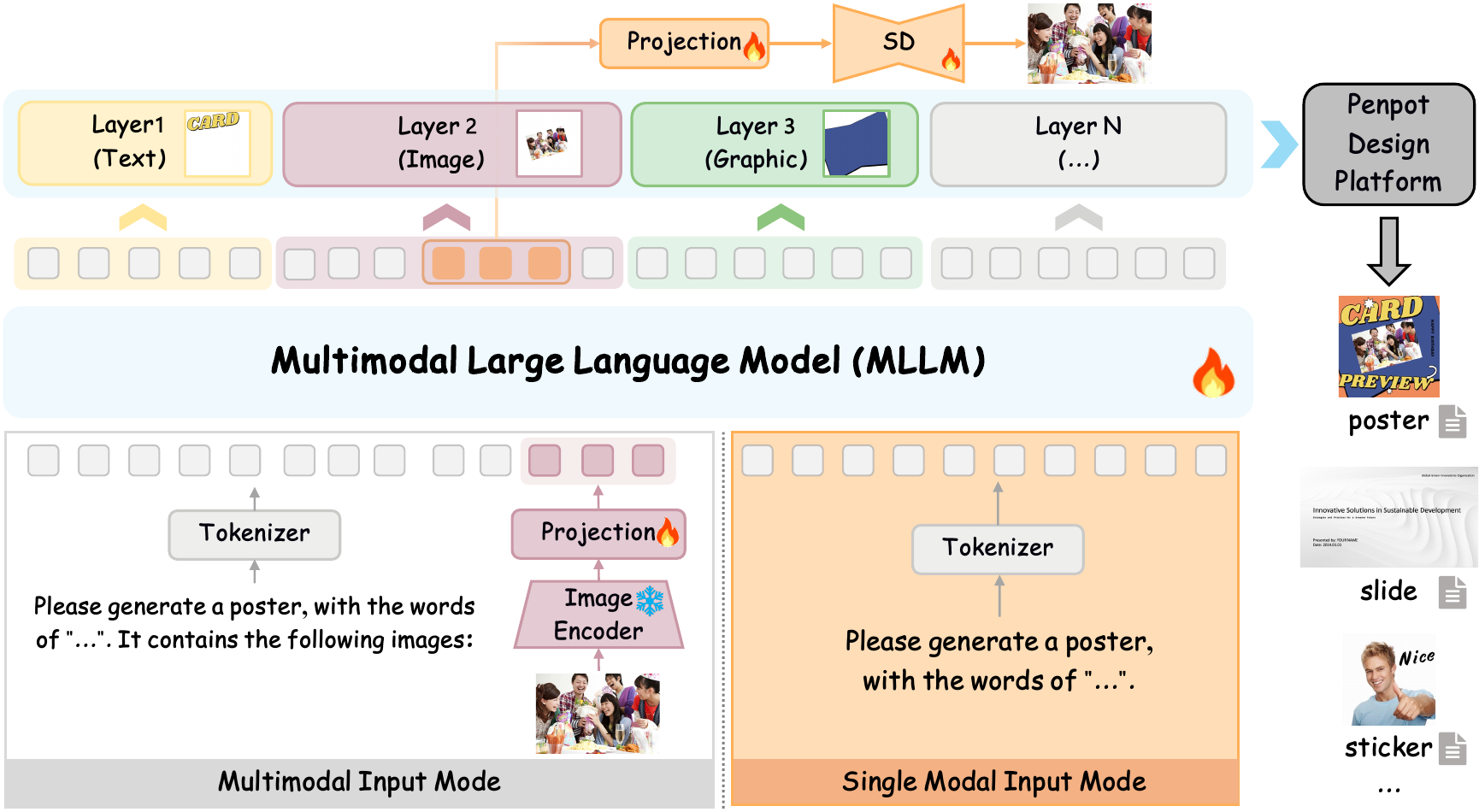}
    \caption{The pipeline of IGD. IGD generates standardized format strings according to instructions to achieve multi-scenario graphic design. In multimodal input mode, IGD receives input from both language and visual modalities to predict elements in language modality. In single modal input mode, IGD only accepts instructions from language and outputs elements from both language and visual modalities.}
    \label{vis:framework}
\end{figure*}

\subsubsection{Data Augmentation}
\label{sec:data_aug}
Given that the graphic design data contains attribute information, we employ random and semantic data augmentation methods to increase diversity. Due to the layout constraints of design files, variations in the character number of text layers can significantly impact visual integrity and compromise data quality. Therefore, we maintain strict consistency in character length for text layers after augmentation.

For random data augmentation, we replace Chinese characters with random characters and English words with random words with the same length to maintain visual uniformity. For semantic augmentation, we use Qwen2.5~\cite{qwen2.5} to generate contextually coherent text, which can retain semantic relevance while introducing unique content.

\subsubsection{Image Processing}
\label{sec:svg}
In graphic designs, simple images such as solid color backgrounds often appear, which allows us to represent these images efficiently. First, we use the k-means clustering method to select the dominant colors to capture its primary color palette. Based on the Potrace algorithm~\cite{potrace}, we transform each dominant color region into a vector path. Combine all the different color paths together into a unified SVG file that captures the color structure. This process achieves scalable and compact representation of simple design elements. Furthermore, for elements that have not been converted to SVG, we use Qwen2-VL~\cite{qwen2vl} to generate a short descriptive tag, which provides contextual information about the design element and enhances controllability.

\subsection{Model Perspective}
\subsubsection{Overview}
As shown in \cref{vis:framework}, IGD supports both multimodal and single modal input mode. Based on the instruction, the model predicts the sequencing order and attributes of each multimodal layer. The standardized string for each generated layer is then sent to Penpot for rendering, resulting in a complete design file. Given that various types of design files share a unified structured format, IGD is broadly applicable across different design scenarios.

\begin{figure}[htbp]
    \centering
    \includegraphics[width=0.82\linewidth]{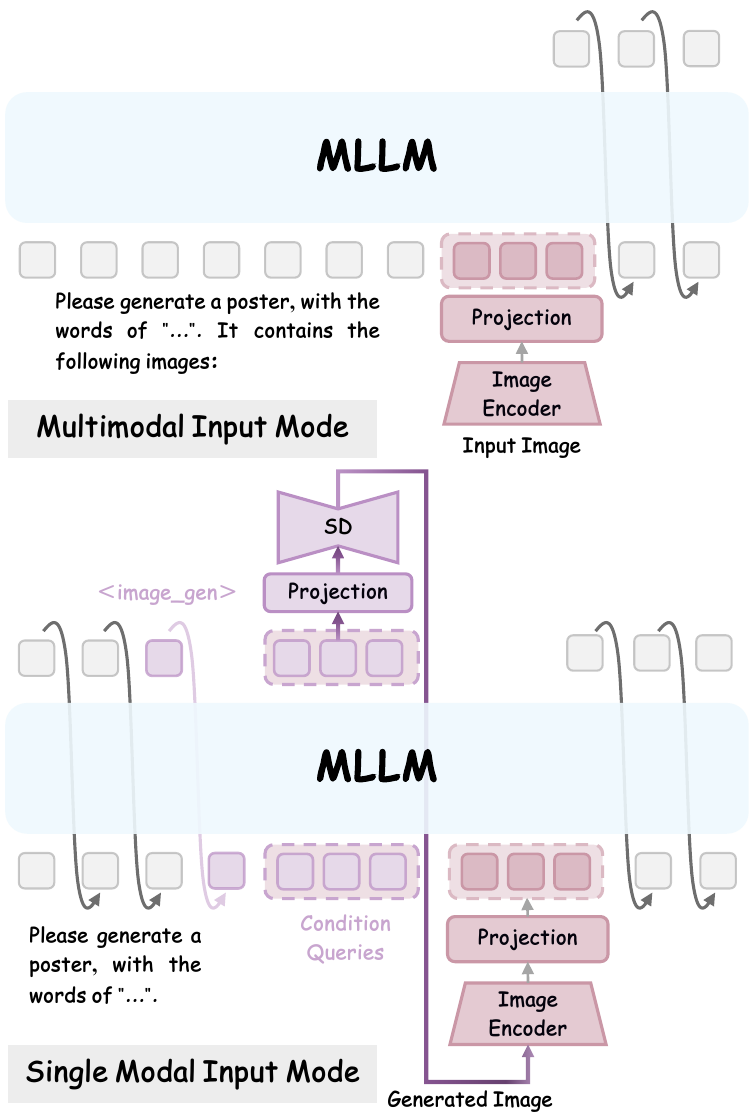}
    \caption{Processes in multimodal and single modal input modes.}
    \label{fig:model_detail}
\end{figure}

\subsubsection{Multi-modal Comprehension and Creation}
As shown in \cref{fig:model_detail}, to enable the synergy between understanding and creation abilities, IGD combines the MLLM $\mathcal{F}$ and diffusion model $\mathcal{H}$. Specifically, LLM adopts Qwen2.5~\cite{qwen2.5} and the diffusion model uses Stable Diffusion (SD)~\cite{stablediffusion}. The features input into SD first go through a simple linear layer $\mathcal{L}_{gen}$. The visual encoder $\mathcal{G}$ incorporates pretrained CLIP~\cite{clip} visual encoder ViT-L/14, followed by a linear layer $\mathcal{L}_{com}$.

The primary goal of IGD is to ensure that design files accurately present the specified text or image content as instructed while enhancing visual appeal. Therefore, we divide IGD into multimodal and single modal input mode. In multimodal input mode, users are expected to provide text and image content and the model only outputs text-based elements. Specifically, input text is tokenized and image content is converted into visual embeddings via the visual encoder $\mathcal{G}$ and the projection layer $\mathcal{L}_{com}$. LLM then autoregressively predicts standardized design text outputs. In single modal input mode, users specify only the text content and the model generates corresponding images and graphics to achieve a cohesive visual result. In this mode, the model first autoregressively generates textual content until it encounters \textit{\textless image\_gen\textgreater}, at which point it pauses the autoregressive process. A set of learnable condition queries are attached to the input to generate embeddings, which is then fed into SD as the conditions for image creation. The generated image is then appended to the input, allowing the model to continue autoregressively predicting subsequent elements.

\subsubsection{Bucket Strategy}
Due to the significant scale variation and aspect ratio changes in graphic design images, and inspired by existing methods~\cite{novelai, sdxl}, we categorize the images into a series of pre-defined buckets based on their width and height. All images within a batch belong to the same bucket, enabling the model to generate images with multiple scales and aspect ratios. When generating an image during the inference process, the appropriate bucket is selected based on the width and height of the image layer predicted by the LLM. The image generation process is then completed using the properties defined within the selected bucket.

\subsubsection{Training Stage}
IGD adopts an end-to-end training process, including cross entropy loss for text and diffusion loss for images. The LLM $\mathcal{F}$ and two projection layers $\mathcal{L}_{gen}$ and $\mathcal{L}_{com}$ follow the DreamLLM~\cite{dreamllm} training process to obtain general understanding and creation capabilities. The further instruction fine-tuning process can be divided into two stages.

\noindent \textbf{Stage 1: Augmented pre-train.} Pre-trained LLM only has the ability to understand and reason logically in natural language and cannot be directly used for generating standardized strings in IGD. We use the random augmented data stated in \cref{sec:data_aug} to inject graphic design related reasoning ability and basic layout skills into the model. Additionally, we employ semantically augmented data to endow the model with the ability to layout and determine the layer order based on semantics. Since these augmented text contents may lack overall semantic relevance to image layers, we adopt the multimodal input mode to enable the model to infer preliminary layout based on text length and to follow instruction-specified content requirements. At this stage, we train only LLM $\mathcal{F}$ and projection layer $\mathcal{L}_{com}$.

\noindent \textbf{Stage 2: End-to-end fine-tuning.} 
After the LLM gains the ability to predict layer attributes and layout, we perform collaboratively training on the LLM, SD, and two projection layers using the source data. The single modal input mode and multimodal input mode are adopted to train separately for different tasks.

\begin{table*}[]
\centering
\renewcommand{\arraystretch}{1.0}
\resizebox{0.9\textwidth}{!}
{
    \begin{tabular}{c|c|ccc|ccccc}
        \hline
        \toprule
        Language & Method & Follow Instructions & Editability & Creativity & Quality & Harmony & Accuracy & Layout & Innovation \\ \midrule
        \multirow{6}{*}{Chinese} & SD3.5~\cite{sd3} & \CheckmarkBold & \XSolidBrush & \CheckmarkBold & 55.4 & 75.6 & 39.0 & 59.6 & 58.0  \\
         & DALL$\cdot$E3~\cite{dalle3} & \CheckmarkBold & \XSolidBrush & \CheckmarkBold & \textbf{81.8} & \textbf{95.4} & \underline{55.0} & \textbf{85.2} & \textbf{86.8}  \\
         & AnyText~\cite{anytext} & \CheckmarkBold & \XSolidBrush & \CheckmarkBold & 49.4 & 61.2 & 46.8 & 49.8 & 47.4  \\
         & CanvaGPT*~\cite{canvagpt} & \XSolidBrush & \CheckmarkBold & \XSolidBrush & 75.4 & \underline{87.6} & 35.6 & \underline{71.0} & 56.8  \\
         & OpenCOLE~\cite{opencole} & \CheckmarkBold & \CheckmarkBold & \CheckmarkBold & 56.0 & 81.4 & 33.6 & 57.8 & 63.0  \\
         & IGD & \CheckmarkBold & \CheckmarkBold & \CheckmarkBold & \underline{79.8} & 82.6 & \textbf{88.4} & 71.4 & \underline{65.8} \\
        \midrule
        \multirow{6}{*}{English} & SD3.5~\cite{sd3} & \CheckmarkBold & \XSolidBrush & \CheckmarkBold & 58.2 & 77.8 & 38.6 & 62.6 & 57.4  \\
         & DALL$\cdot$E3~\cite{dalle3} & \CheckmarkBold & \XSolidBrush & \CheckmarkBold & \underline{81.0} & \textbf{96.2} & 52.8 & \textbf{82.4} & \textbf{88.6}  \\
         & AnyText~\cite{anytext} & \CheckmarkBold & \XSolidBrush & \CheckmarkBold & 37.6 & 54.2 & 30.2 & 37.2 & 42.2  \\
         & CanvaGPT*~\cite{canvagpt} & \XSolidBrush & \CheckmarkBold & \XSolidBrush & 76.8 & 86.4 & 54.0 & 70.8 & 59.6 \\
         & OpenCOLE~\cite{opencole} & \CheckmarkBold & \CheckmarkBold & \CheckmarkBold & 70.4 & 85.2 & \underline{65.6} & 62.2 & 66.0  \\
         & IGD & \CheckmarkBold & \CheckmarkBold & \CheckmarkBold & \textbf{81.2} & \underline{88.8} & \textbf{81.4} & \underline{72.0} & \underline{68.2} \\
        \midrule
        \multicolumn{2}{c|}{Training Samples} & - & - & - & 88.8 & 89.2 & 93.6 & 82.6 & 70.4 \\
        \bottomrule
    \end{tabular}
}
\caption{GPT evaluation scores. ``Follow Instruction'' indicates whether the text is consistent with the information intended to be conveyed. ``Editability'' means whether design files allow for editing. ``Creativity'' reflects the capability for diverse content generation. Due to CanvaGPT not providing an API, we manually generate 100 evaluation samples. We also take 1k training samples for the GPT evaluation.}
\label{tab:compare_table}
\end{table*}

\subsection{Discussion}
\textbf{Why convert simple images to SVG?} Unlike images in typical text-to-image tasks, which carry complete semantics and distinct object targets, graphic design images are often composed of solid colors or simple abstract patterns, generally lacking meaningful semantic content. On the one hand, using image generation methods for these simple graphics significantly increases model complexity and thus reduces the learning efficiency. Converting these simple images to SVG requires only a small number of text tokens, effectively representing them without inflating token number or introducing another modality. On the other hand, due to the large volume and similar pattern of simple images, the data exhibits a long-tail distribution. During training, the model tends to favor simple colors and patterns, which may result in ignoring complex details and weakening the learning of primary content targets.

\noindent \textbf{Why perform end-to-end training for LLM and SD?} Using the LLM to generate natural language descriptions and feeding them directly into SD has the following limitations: 1) Constraints of natural language descriptions. Natural language alone struggles to capture the fine details of an image, often leading to subjective and potentially biased descriptions, which can introduce ambiguity and imprecision in representing fine-grained visual characteristics. 2) Lack of inter-layer correlation. When generating images, the model needs to account for the spatial relationships and stylistic coherence between layers. Since the image generation supervision cannot feed back into the LLM, relying solely on natural language fails to convey essential inter-layer relationships, reducing overall consistency across layers. 3) Inability to conduct end-to-end training. All images in graphic design files require additional captions to separately train both models, which not only adds inconvenience but also introduce potential gaps in image captioning.

\section{Experiments}
\begin{table}[]
\centering
\renewcommand{\arraystretch}{1.0}
\resizebox{\linewidth}{!}
{
    \begin{tabular}{c|ccc|ccc|c}
        \hline
        \toprule
        Method & Char-P $\uparrow$ & Char-R $\uparrow$ & Char-F $\uparrow$ & $R_{ali} \downarrow$  & $R_{ove} \downarrow$ & $R_{com} \downarrow$ & FID $\downarrow$ \\
        \midrule
        SD3.5~\cite{sd3} & 70.55 & 77.43 & 73.83 & - & - & - & 107.42 \\
        DALL$\cdot$E3~\cite{dalle3} & 37.08 & 50.64 & 42.81 & - & - & - & 159.37 \\
        AnyText~\cite{anytext} & 83.33 & 38.04 & 52.23 & - & - & - & 240.36 \\
        OpenCOLE~\cite{opencole} & 58.21 & 59.31 & 58.76 & 0.1941 & 0.0347 & 25.18 & 122.20 \\
        IGD & \textbf{94.35} & \textbf{80.41} & \textbf{86.83} & \textbf{0.0541} & \textbf{0.0245} & \textbf{21.53} & \textbf{56.35} \\
        \midrule
        OpenCOLE$\dag$~\cite{opencole} & 69.46 & 85.42 & 76.62 & - & - & - & - \\
        IGD$\dag$ & \textbf{98.19} & \textbf{96.60} & \textbf{97.39} & - & - & - & - \\
        \bottomrule
    \end{tabular}
}
\caption{More metrics with English user inputs. $\dag$ indicates the calculation of text accuracy using predicted text layer raw typographic information, i.e. before rendering into an image.}
\label{tab:compare_table2}
\end{table}

\begin{figure*}[htbp]
    \centering
    \includegraphics[width=0.95\textwidth]{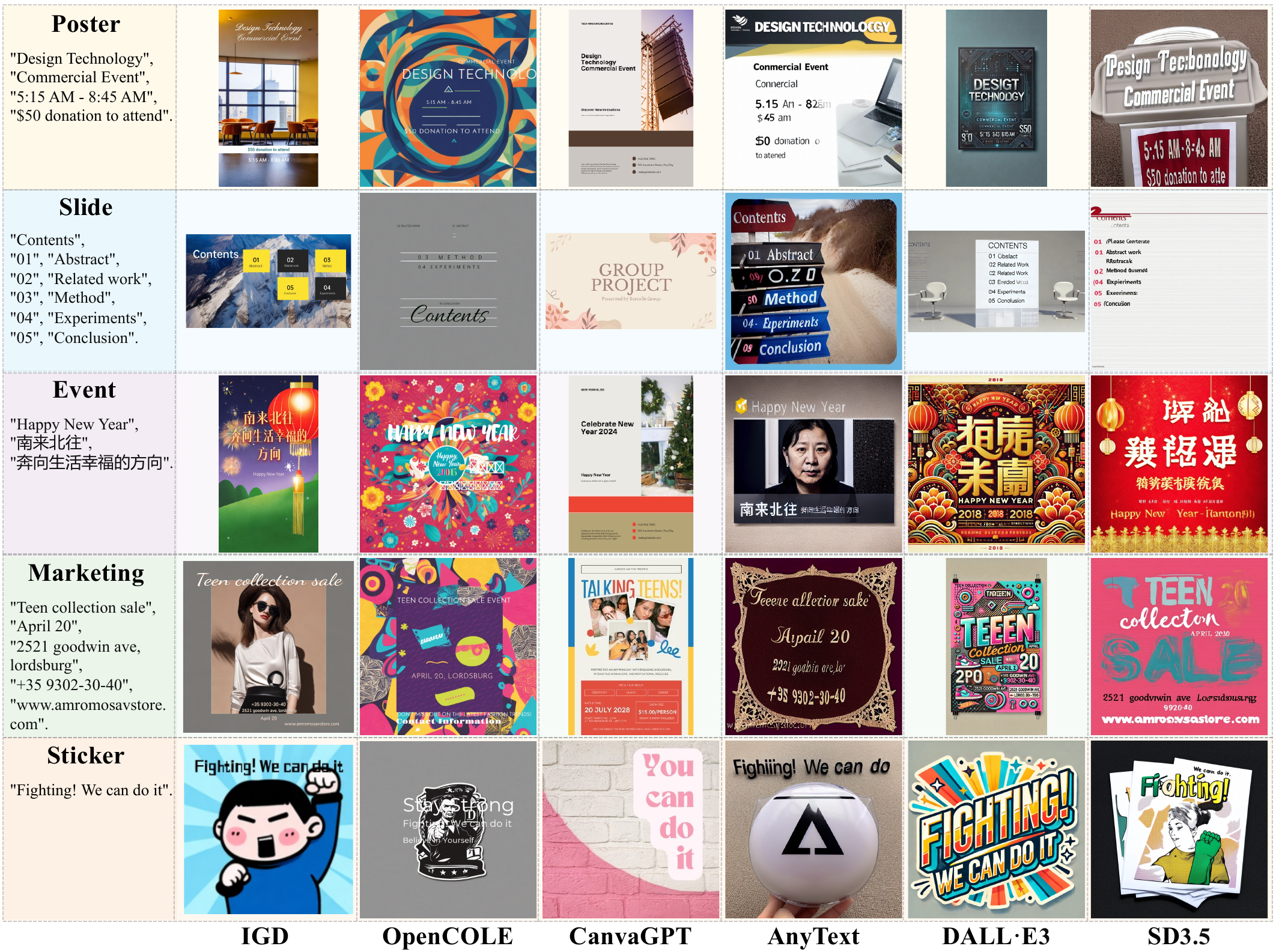}
    \caption{Comparison with the state-of-the-art methods. IGD effectively adheres to instructions by presenting the intended content and generating corresponding types of graphic design files. While maintaining creativity, it also provides editorial flexibility through the generation of graphic layers. Zoom in for a better view.}
    \label{fig:compare_vis}
\end{figure*}

\subsection{Datasets and Evaluation Metrics}
We collected 90k training samples, including free data from the Internet and paid data. 1k each of English and Chinese user inputs are prepared for evaluation. We utilized GPT-4o for evaluation across five dimensions: Quality, Harmony, Accuracy, Layout and Innovation. We also evaluate text accuracy by calculating character-level precision, recall, and f-measure from OCR results. For graphic metrics, we adopt $R_{ali}$, $R_{ove}$~\cite{arroyo2021variational}, and $R_{com}$~\cite{cglgan} to evaluate the alignment between multimodal layers, overlap between text layers, and the gradient of pixels in the text layer regions. For more details, please refer to the supplementary materials.

\begin{figure*}[htbp]
    \centering
    \includegraphics[width=0.96\textwidth]{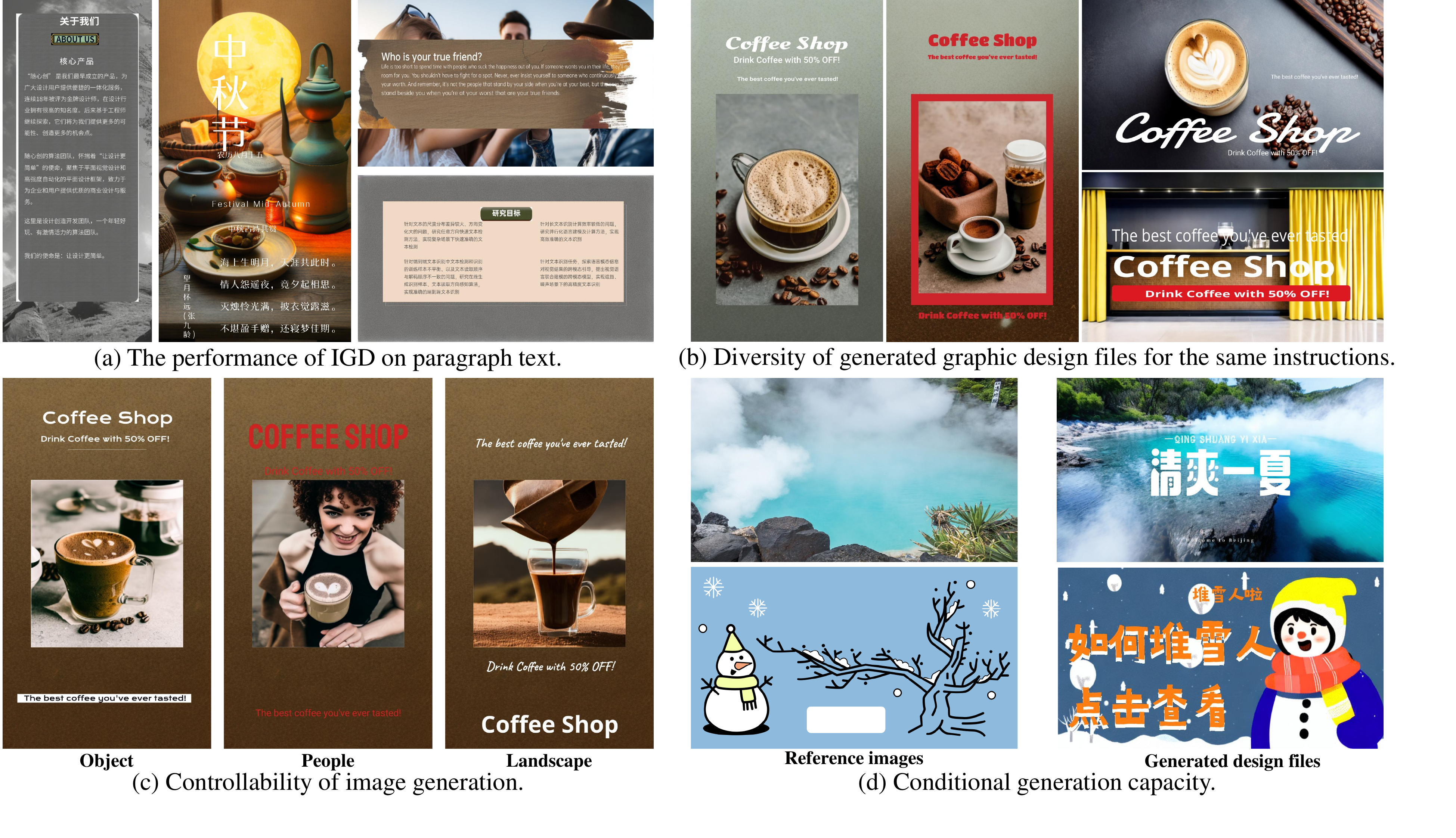}
    \caption{Several samples to demonstrate the characteristics of IGD.}
    \label{fig:all-discuss}
\end{figure*}

\subsection{Comparison with Existing Methods}
\cref{tab:compare_table} and \cref{fig:compare_vis} show the qualitative and quantitative comparison results between IGD and existing methods. As the diffusion-based methods~\cite{anytext, dalle3, sd3} generate design images as a whole, they exhibit strong harmony. However, they are limited to image-level generation and cannot accommodate diverse graphic design types. Visual text generated using diffusion models also suffers from repetition, omission, distortion, and spelling errors. Additionally, due to the methodological characteristics, the generated images lack interactive flexibility, making these methods difficult to apply in practical scenarios. CanvaGPT~\cite{canvagpt} performs well in terms of layer consistency and typography, benefiting from its combination of GPT and well-designed templates. However, the quantity and quality of these templates greatly limit their performance. When confronted with diverse and personalized instructions, the generated design files diverge substantially from the information intended to convey, thus compromising the accuracy and effectiveness of the message. OpenCOLE~\cite{opencole} possesses various basic capabilities, but it only supports generating a single background image and can only predict text layers with a few attributes. OpenCOLE does not support Chinese rendering and is limited in handling multi-scene, multi-resolution design images. In comparison, IGD offers stronger versatility, supporting both Chinese and English. Additionally, due to the end-to-end optimization process, IGD generates more consistent and holistic design images. IGD retains the creativity of the diffusion model while adapting to various types of graphic design scenarios, benefiting from the paradigm of parameter prediction and rendering.

The OCR results in \cref{tab:compare_table2} further confirm that IGD maintains high accuracy in rendered text (86.83\% vs 73.83\% for SD3.5~\cite{sd3}). Due to potential layer overlaps during rendering and limitations of the OCR model, the F-measure of text before rendering is 20.77\% higher than OpenCOLE. For graphic metrics, a lower $R_{ali}$ indicates that IGD performs better in layer alignment, while the overlap $R_{ove}$ between text layers and the gradient $R_{com}$ within text layers demonstrate that IGD generates text with less interference, resulting in higher readability. The FID metric for IGD is far superior to the other methods (56.35 vs 107.42 for SD3.5). It is worth noting that the Inception model is trained on generic datasets rather than design-specific datasets. As a result, the extracted feature may not fully capture the unique characteristics of the design graphic.

\subsection{Ablation Study}
We conduct several ablation experiments to verify the impact of critical components on model performance. For data augmentation, omitting the augmented pre-training in stage 1 leads to a decline in layout performance and overall image harmony. Although the augmented data lacks semantic coherence or association with other multimodal layers, they enable the model to have reasoning capabilities related to graphic design and determine the layout according to the text length and basic semantics, thereby enhancing holistic performance. Regarding image processing, converting solid-color and simple abstract pattern images into SVG format for parametric generation reduces training complexity. The model trained without such conversion tend to generate simple and flat images that lack clear semantics and primary objects. The foundational reasoning capability of LLM is also significant. Substituting Vicuna for Qwen2.5 leads to a decline in layout arrangement ability, subsequently impairing aesthetic quality and text readability in design images. Finally, integrating LLM and SD in an end-to-end training process improves spatial relationships and stylistic consistency across multimodal layers, ensuring harmonious compositions. Freezing SD parameters during training slightly compromises overall performance but achieves marginally better results in layer alignment and overlap. See the supplementary materials for details.

\subsection{Characteristic Analysis}
\noindent \textbf{Large Paragraphs of Text} \cref{fig:all-discuss}(a) illustrates the ability to generate layouts for paragraph text and dense text. The results show that IGD exhibits strong typographic capabilities across various text types. Owing to the attribute prediction and parameter rendering paradigm, all text within graphic design images is clear and legible. Furthermore, IGD demonstrates robust instruction adherence, accurately conveying the intended information even with extensive text content, validating its effectiveness and practicality.

\noindent \textbf{Diversity} Creativity and diversity in graphic design are crucial elements. \cref{fig:all-discuss}(b) show multiple randomly generated results based on the same input instructions, demonstrating that IGD achieves notable diversity in visual elements, layout structures, and design styles. This versatility provides users with insightful ideas and inspiration for their projects.

\noindent \textbf{Controllability of Image Generation} When IGD stops at \textit{\textless image\_gen\textgreater} to generate the subsequent image, we modify the description tag of the previous image in \textit{\textless image\_des\textgreater}, which is stated in \cref{sec:svg}. As shown in \cref{fig:all-discuss}(c), IGD shows promising controllability when we force the description of the foreground image from \textit{object} to \textit{person} and \textit{landscape}.

\noindent \textbf{Conditional Generation Capacity} As illustrated in \cref{fig:all-discuss}(d), when a model trained in the single modal input mode is provided with both text and image content for inference, it exhibits a strong ability to imitate the input image style. We attribute this to the training procedure of the single modal input mode, which also incorporates multimodal understanding, and the inter-layer consistency across multimodal layers brought by the end-to-end synergistic training of LLM and SD. It demonstrates that IGD can effectively capture and perceive preceding multimodal information.

\section{Conclusion} 
We investigate the primary challenges in graphic design and take a further step towards automated graphic design. On the data aspect, we develope the Penpot design platform and establish a standardized format, achieving unified representation of graphic design files across multiple scenarios, thereby laying the foundation for scalable data. On the model aspect, we propose the Instructional Graphic Designer based on the unified framework for multimodal understanding and creation. Compared to existing methods, IGD effectively follows instructional requirements and accurately conveys the intended information and data with satisfactory layouts. Benefiting from the development of the Penpot design platform and multimodal layer elements, the generated design files offer flexibility for interactive editing. In quantitative metrics and qualitative analyses, IGD demonstrates superior performance and great potential, providing a new solution for fully automated graphic design.

\section*{Acknowledgments}
This work is supported by the the National Nature Science Foundation of China (62425114, 62121002, U23B2028, 62232006). We acknowledge the support of GPU cluster built by MCC Lab of Information Science and Technology Institution, USTC. We also thank the USTC supercomputing center for providing computational resources.

{
    \small
    \bibliographystyle{ieeenat_fullname}
    \bibliography{main}
}

\end{document}